\documentclass[letterpaper, 10 pt, conference]{ieeeconf}  

\IEEEoverridecommandlockouts                              

\usepackage{amsmath,amsfonts}
\usepackage{algorithm}
\usepackage{algpseudocode}
\usepackage{array}
\usepackage[caption=false,font=normalsize,labelfont=sf,textfont=sf]{subfig}
\usepackage{textcomp}
\usepackage{stfloats}
\usepackage{url}
\usepackage{verbatim}
\usepackage{graphicx}
\usepackage{cite} 
\usepackage{xcolor}

\title{\LARGE \bf
A Grasp Pose is All You Need: Learning Multi-fingered Grasping with Deep Reinforcement Learning from Vision and Touch
}

\author{Federico~Ceola,
        Elisa~Maiettini,
        Lorenzo~Rosasco
        and~Lorenzo~Natale
\thanks{Federico Ceola, Elisa Maiettini and Lorenzo Natale are with Humanoid Sensing and Perception (HSP), Istituto Italiano di Tecnologia (IIT), Genoa, Italy (email: {\tt\footnotesize name.surname@iit.it}).}
\thanks{Federico Ceola and Lorenzo Rosasco are with Laboratory for Computational and Statistical Learning (LCSL), with Machine Learning Genoa Center (MaLGa) and with Dipartimento di Informatica, Bioingegneria, Robotica e Ingegneria dei Sistemi (DIBRIS), University of Genoa, Genoa, Italy.}
\thanks{Lorenzo Rosasco is also with Istituto Italiano di Tecnologia (IIT) and with Center for Brains, Minds and Machines (CBMM), Massachusetts Institute of Technology (MIT), Cambridge, MA (email: {\tt\footnotesize lrosasco@mit.edu}).}
}

\begin{document}

\maketitle
\thispagestyle{empty}
\pagestyle{empty}


\begin{abstract}

Multi-fingered robotic hands have potential to enable robots to perform sophisticated manipulation tasks. However, teaching a robot to grasp objects with an anthropomorphic hand is an arduous problem due to the high dimensionality of state and action spaces.
Deep Reinforcement Learning (DRL) offers techniques to design control policies for this kind of problems without explicit environment or hand modeling. 
However, state-of-the-art model-free algorithms have proven inefficient for learning such policies.
The main problem is that the exploration of the environment is unfeasible for such high-dimensional problems, thus hampering the initial phases of policy optimization.
One possibility to address this is to rely on off-line task demonstrations, but, oftentimes, this is too demanding in terms of time and computational resources.

To address these problems, we propose the \textit{A Grasp Pose is All You Need} (G-PAYN) method for the anthropomorphic hand of the iCub humanoid. We develop an approach to automatically collect task demonstrations to initialize the training of the policy. The proposed grasping pipeline starts from a grasp pose generated by an external algorithm, used to initiate the movement. Then a control policy (previously trained with the proposed G-PAYN) is used to reach and grab the object. We deployed the iCub into the MuJoCo simulator and use it to test our approach with objects from the YCB-Video dataset. Results show that G-PAYN outperforms current DRL techniques in the considered setting in terms of success rate and execution time with respect to the baselines.

The code to reproduce the experiments is released together with the paper with an open source license\footnote{\url{https://github.com/hsp-iit/rl-icub-dexterous-manipulation}}.

\end{abstract}


\section{INTRODUCTION}
\label{sec:introduction}
Robotic grasping is one of the most important manipulation tasks, due to its importance for other downstream tasks such as pick-and-place~\cite{7583659}, in-hand manipulation~\cite{chen2022system}, or objects stacking~\cite{lee2021beyond}.

While two-fingered grasping has been extensively studied in the literature~\cite{Mahler-RSS-17, gpd, levine2018learning, 9216986}, grasping with multi-fingered robotic hands is still an open problem. Although grasping with two-fingered grippers is easier to plan and execute, anthropomorphic hands offer the opportunity to perform dexterous tasks such as objects re-orientation~\cite{openai2018learning}, and enable robots to use tools such as hammers~\cite{Rajeswaran-RSS-18}. However, due to the intrinsic difficulty of the task, which requires controlling tens of degrees of freedom (DoFs) finding suitable manipulation strategies is challenging.

The latest advancements in the DRL literature provide tools to design high-dimensional control policies without requiring specific environment and hand modeling. 
State-of-the-art model-free algorithms such as SAC~\cite{haarnoja2018soft} or PPO~\cite{schulman2017proximal}, have proven inefficient to learn policies on multi-fingered manipulation tasks. This is due to the fact that, in these cases, an efficient exploration of the environment at the beginning of policies optimization is unfeasible due to the high dimensionality of the problem. Some recent methods propose to address this problem leveraging on data acquired from off-line task demonstrations, and to combine them with data acquired on-line during policy training. While these approaches have shown promising results, collection of demonstrations is a non-trivial procedure which requires appropriate tools, such as MoCap~\cite{taheri2020grab} or Virtual Reality~\cite{7363441} systems.

Another problem with state-of-the-art methods is that they typically use information such as joints and objects poses and velocities that are difficult to obtain, or that can be noisy in practice~\cite{Rajeswaran-RSS-18, dasari2022learning, kumar2019contextual}.

\begin{figure}
    \vspace{2mm}
    \centering
    \includegraphics[width=0.92\linewidth]{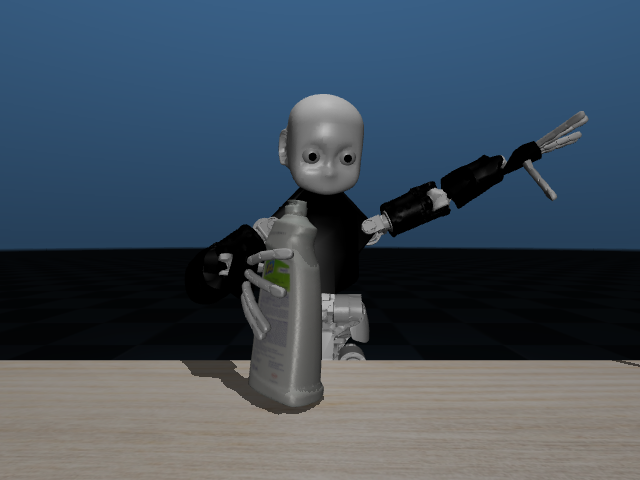}
    \caption{\textbf{iCub simulated environment}.}
    \label{fig:icub_sim}
\end{figure}

In this work we we aim at overcoming these limitations by proposing a grasping method for the iCub~\cite{icub} humanoid robot based on DRL that leverages on automatically collected demonstrations. To our knowledge, this is the first method that learns this task from RGB data, tactile and proprioceptive information. We start from a grasp pose generated by an external algorithm, using it as a prior information for our task. 
We assume that this initial pose is inaccurate and should be refined given the specific object and grasping hand. For this reason, 
the robot first moves the end-effector close to this pose to start the grasping movement, and then uses a separate policy to approach and grasp the object. We train the policy with the proposed \textbf{G-PAYN}. The method firstly automatically acquires a set of demonstrations leveraging on a given grasp planner, and then it trains a policy starting from the data originating from the execution of such demonstrations. We design a reward function for the training process that  uses a measure of grasp success or failure, but also takes into consideration intermediate steps of the grasping movement. For example, we use information from the tactile sensors, and a positive reward for those hands configurations that increase the number of contacts to achieve a more stable grasp.

We test our approach on five objects from the YCB-Video~\cite{xiang2018posecnn} dataset and we consider two different grasp pose generators to evaluate how their choice affects performance. We benchmark our method against three DRL baselines, outperforming them in all the experiments. Moreover, we compare the success rate of the proposed grasping pipeline against the baseline used to generate the task demonstrations in the proposed \textbf{G-PAYN}. 
Experiments show that the learned policies surpass the baseline in half of the cases and perform comparably well in the remaining instances. This demonstrates that the proposed \textbf{G-PAYN} does not just imitate the behavior of the off-line demonstrations but it refines the movements, adapting them to the specific object.

We run our experiments in a simulated tabletop setting. To this aim, we deployed a MuJoCo~\cite{todorov2012mujoco} model for the iCub humanoid, which, as a further contribution of this paper, we also make publicly available together with the code to reproduce the experiments.
In Fig.~\ref{fig:icub_sim}, we show a snapshot of the simulated environment.

\begin{figure*}
    \vspace{2mm}
    \centering
    \includegraphics[width=0.94\linewidth]{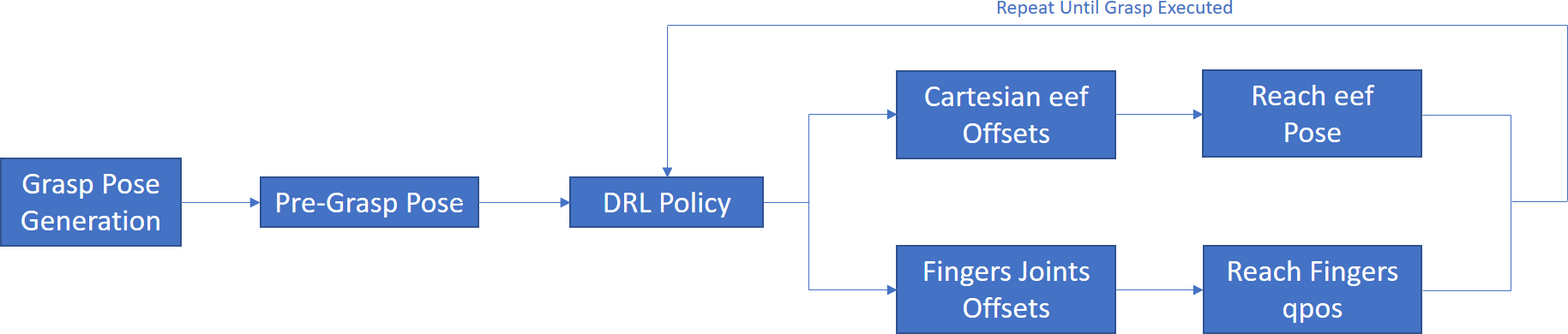}
    \caption{\textbf{Overview of the proposed grasping pipeline}. We rely on a \textit{Grasp Pose Generator} to compute a suitable grasp pose for the considered object. Then, we move the end-effector of the robot to a \textit{Pre-Grasp Pose} close to the previously generated grasp pose. Finally, we use a \textit{DRL Policy} to predict cartesian offsets to move the end-effector toward the object and offsets in the joint space of the fingers to grasp it. We repeat this procedure until the grasp is executed.}
    \label{fig:pipeline}
\end{figure*}

Finally, in the video attached as supplementary material to the manuscript\footnote{\label{video_fn}\url{https://youtu.be/qc6gksKH3Mo}}, we show a grasping demonstration on the real iCub humanoid, relying on a policy trained in simulation.

\section{RELATED WORK}
\label{sec:relwork}
We propose a DRL-based application for multi-fingered grasping, leveraging on automatically collected demonstrations to train our policy. In the following subsections, we cover the main literature on multi-fingered grasping and on DRL from demonstrations.

\subsection{Multi-fingered Grasping}

Multi-fingered grasping is a challenging task due to the high number of DoFs involved and complex hand-object interactions. Some recent works~\cite{multifinger, li2022efficientgrasp} propose methods for multi-fingered grasp synthesis starting from pointcloud information. These methods are difficult to apply because they are constrained to the hardware that is used for training, and they do not take into account the hand-object interactions during grasp execution. The approach described in~\cite{liang2021multifingered_rl} deals with the high number of DoFs in the Shadow hand with a PCA-based hand synergy. Then, similarly to our solution, it trains a DRL policy to grasp an object starting from a grasp pose given by an external algorithm. However, this method uses as input to the policy binary tactile information, joint torques (which might not be available in all the robots) and hand joints positions, without considering any information of the object (e.g. object position or visual feedback) that would allow grasp recovery if the grasp pose is not suitable. Other approaches, such as the one in~\cite{chen2022dextransfer}, instead learn grasping policies using data collected with a MoCap system, with the aim of reducing the amount of training data, since the data collection procedure for multi-fingered grasping is challenging. In our approach, we propose a method to overcome this issue by automatically collecting off-line demonstrations.

\subsection{Deep Reinforcement Learning from Demonstrations}
Methods that learn DRL policies leveraging on demonstrations can be grouped in two categories. The first is composed of methods that use demonstrations throughout the training. Two exemplar methods are DDPGfD~\cite{vecerik2017leveraging} and the approach proposed in~\cite{nair2018overcoming}. Both methods modify the DDPG~\cite{lillicrap2015continuous} algorithm to leverage on the demonstrations included in the replay buffer. The second class of methods uses the demonstrations for pre-training a policy, either with behavior cloning or with DRL, and then fine-tunes such policy on data acquired on-line. Two exemplar approaches are DAPG~\cite{Rajeswaran-RSS-18} and AWAC~\cite{nair2020awac}. The former learns several dexterous manipulation tasks, leveraging on pre-training from demonstrations with behavior cloning and then fine-tuning the policy with an augmented loss to stay close to the demonstrations. The latter mitigates the distribution shift between the off-line demonstrations and the data acquired on-line during training. In this work, we consider an approach from each of the above classes as baselines for our experiments, namely \cite{nair2018overcoming}, adapted to the considered setting, and AWAC~\cite{nair2020awac}.

\section{METHODOLOGY}
\label{sec:methods}

\begin{figure}
    \vspace{2mm}
    \centering
    \includegraphics[width=0.94\linewidth]{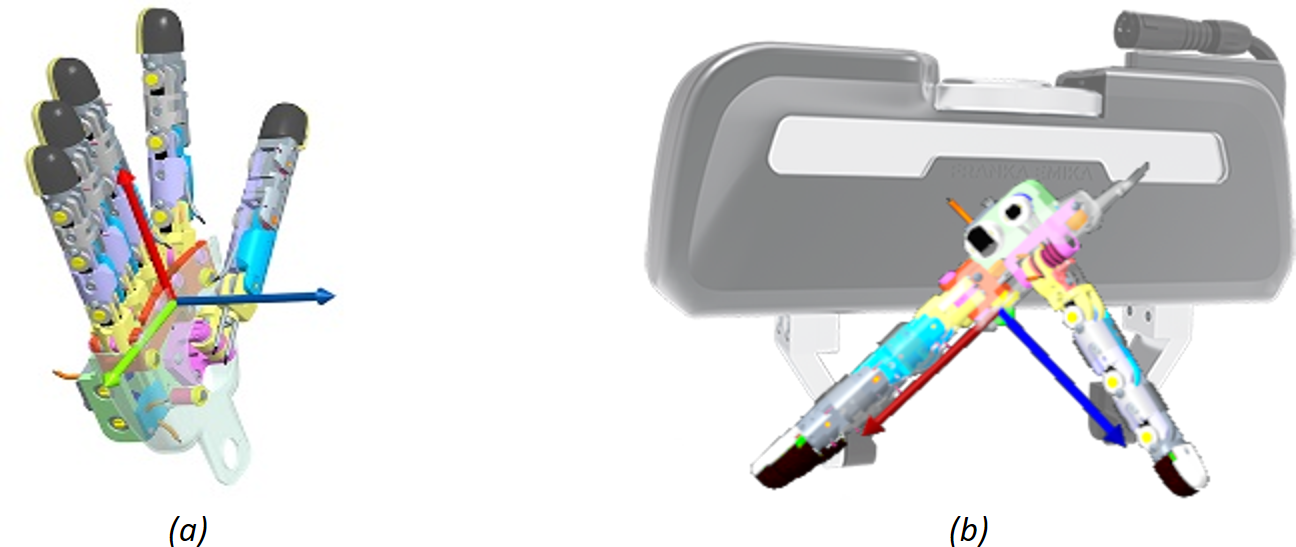}
    \caption{\textit{(a)} \textbf{iCub hand reference frame.} \textit{(b)} \textbf{VGN grasp transformation for the iCub hand.} We rotate the grasp pose generated for the Franka Emika Panda gripper by $45$° to obtain the corresponding grasp pose for the iCub hand.}
    \label{fig:hand}
\end{figure}

\subsection{Grasping Pipeline}
\label{sec:grasp_pipeline}

We propose a modular pipeline for grasping an object with the iCub humanoid (see Fig.~\ref{fig:pipeline}). Our approach is composed of two stages. We firstly compute a suitable grasp pose with an external algorithm and we move the end-effector of the robot in a pre-grasp pose spaced $5 cm$ from the one given by the algorithm. Then, starting from this configuration, we rely on a DRL policy to move the end-effector in the cartesian space and to control the position of the fingers joints in order to accomplish the grasp and lift the object.

\subsubsection{Grasp pose computation} we rely on two different algorithms for grasp pose computation:

\begin{itemize}
    \item The approach based on superquadric models proposed in~\cite{vezzani2017sq}. This is specifically designed to compute a grasp pose for the iCub humanoid.
    \item The state-of-the-art grasp pose generator VGN~\cite{breyer2020volumetric}. This approach computes grasp poses for a two-fingered gripper mounted on a Franka Emika Panda robot. To compute a feasible grasp pose for the iCub, we post-process the grasping candidates proposed by VGN by rotating the original grasp pose of $45$°, see Fig.~\ref{fig:hand} \textit{(b)}. We then analyze the reachability of the rotated grasp poses in decreasing order of confidence until a suitable candidate is found.
\end{itemize}

\subsubsection{Grasp execution} 
\label{subsec:grasp_exec}
we model the grasp execution task as a Markov Decision Process (MDP) $\{S, A, T, r\}$. At each timestep $t$, the robot observes the state $s_t \in S$ of the system. This has five components:

\begin{itemize}
    \item A visual information of the environment. This is computed starting from the RGB image acquired by the camera mounted on the iCub's head at a given timestep $t$. We process this with the \textit{ViT-B/32} model\footnote{We rely on the pre-trained models available at \url{https://github.com/openai/CLIP}.} pre-trained with CLIP~\cite{radford2021learning} to extract a latent representation of the image. Finally, to encode temporal information, we combine this latter with the latent representations obtained at the timesteps $t-1$ and $t-2$ with the \textit{Flare} architecture~\cite{shang2021reinforcement}.
    \item The cartesian pose of the end-effector.
    \item Fingers joints position (\textit{qpos}).
    \item A binary tactile value for each of the fingertips of the iCub hand, encoding the information of the contact with the object to grasp.
    \item The center of the superquadric that fits the object to be grasped, or the median point of object's pointcloud for VGN. This value is taken at the beginning of the episode and remains unchanged throughout its duration. We estimate the center of the object in this way because on the real robot the actual position of the object is not available. This estimate may be error prone and non-deterministic (e.g. due to the random sampling of the pointcloud for superquadrics estimation), and therefore it may affect the trained policies. However, we show that our approach is able to learn grasping policies using the considered, possibly imprecise, estimation of the center of the object.
\end{itemize}

At each timestep $t$, given the current state of the system $s_t$, the robot acts in the environment with unknown dynamics $T$, according to a policy $\pi (a_t | s_t)$. Specifically, the action $a_t \in A$ has two components:

\begin{itemize}
    \item A cartesian offset that represents the movement that the end-effector of the robot needs to accomplish w.r.t. its current cartesian pose. This is used to reach the object in a suitable configuration, starting from the pre-grasp pose.
    \item The movement that the fingers joints have to perform w.r.t. the current \textit{qpos}. This is needed to close the fingers and grasp the object.
\end{itemize}

We learn the policy $\pi^*$ with the SAC~\cite{haarnoja2018soft} algorithm by maximizing the maximum entropy objective:

\begin{align}
    \label{eq:max_ent}
    \pi^* = \arg\max_{\pi} \sum_{t} \mathbb{E}_{(s_t, a_t) \sim \rho_\pi}[r(s_t, a_t) + \alpha\mathcal{H}(\pi(\cdot|s_t))],
\end{align}

This adds a weighted entropy component $\alpha\mathcal{H}(\pi(\cdot|s_t))$ to the standard action-value function $\arg\max_{\pi} \sum_{t} \mathbb{E}_{(s_t, a_t) \sim \rho_\pi}[r(s_t, a_t)]$, where $\rho_\pi$ denotes the state-action marginal of the trajectory distribution induced by a policy $\pi$, with the aim of encouraging exploration during training. Specifically, for our task, we design the reward function $r(s_t, a_t)$ in eq. (\ref{eq:max_ent}), which evaluates the outcome of action $a_t$ when the state of the environment is $s_t$, as the sum of the following components:

\begin{figure*}
    \vspace{2mm}
    \centering
    \includegraphics[width=0.98\linewidth]{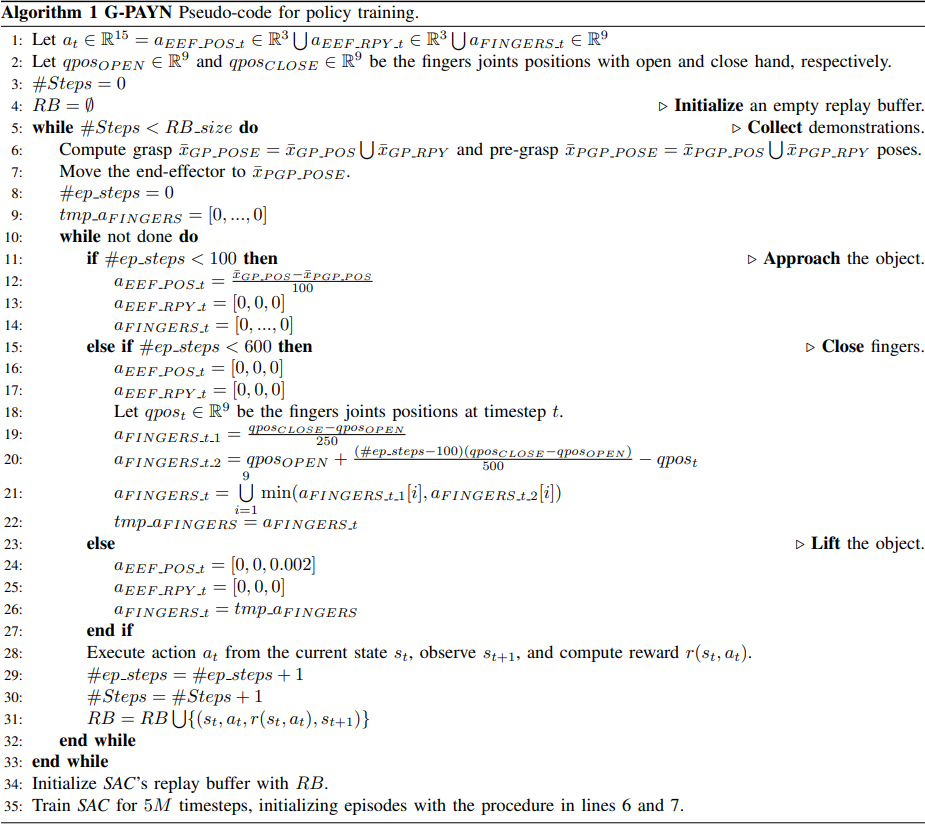}\ContinuedFloat
    \label{fig:my_label}
\end{figure*}

\begin{itemize}
    \item $r_{fingers} = f(t+1) - f(t)$, where $f$ denotes the number of fingers in contact with the object at the given timestep.
    \item $r_{dist\_object\_center}$ guides the end-effector to approach the object. It considers the distance $d$, measured in $cm$, of the object position estimated at the beginning of the episode w.r.t. the $x$ and $y$ axes of the iCub hand reference frame (red and green axes in Fig.~\ref{fig:hand} \textit{(a)}). Specifically, for the superquadrics-based approach, we consider the center of the superquadric, while we use the median point of the object's pointcloud, when considering VGN.
    The value of this component is equal to $d(t+1) - d(t)$ if $f(i) < 2$ for all the timesteps $i \in {0,...,t+1}$, otherwise it is equal to $0$.
    \item $r_{object\_height}$: this component rewards the difference of the object height $h$ measured in $mm$ with respect to the table between two consecutive timesteps, $t+1$ and $t$, therefore evaluating whether the object has been lifted from the table in that time frame. We consider this term only from the moment when at least two fingers touch the object. Specifically, if $f(t+1) \geq 2$ and  $h(t+1) - h(t) > 0$, or $f(i) \geq 2$ for at least one timestep $i \in {0,...,t+1}$ and $h(t+1) - h(t) < 0$, $r_{object\_height} = f(t+1)*(h(t+1) - h(t))$, otherwise $r_{object\_height} = 0$.
    \item $r_{end \_of\_episode}$ evaluates the termination of the episode. An episode might end in one of the following cases: \textit{(i)} the object is grasped and uplifted by $10 cm$, \textit{(ii)} the object is moved too far from the initial position, \textit{(iii)} the inverse kinematic solver cannot find a solution for the required configuration of the end-effector, or \textit{(iv)} the current timestep is equal to the maximum number of allowed timesteps per episode. In the first case, the robot has successfully grasped the object and the term $r_{end \_of\_episode}$ is equal to $1$. In all other cases, the episode is considered a failure and thus the term is equal to $-1$.
\end{itemize}

\subsection{Policy Training}
\label{subsec:policy_training}
We propose a two-stage algorithm to learn the grasping policy described in Sec.~\ref{sec:grasp_pipeline}: we first design an automatic procedure for the acquisition of grasping demonstrations, and then we learn the policy with the SAC~\cite{haarnoja2018soft} algorithm, leveraging on the previously acquired data.

The proposed pipeline for demonstrations collection is composed of three steps. Firstly, once the end-effector has reached the considered pre-grasp pose (as described in Sec.~\ref{sec:grasp_pipeline}), we move it toward the grasp pose on a straight line by splitting the trajectory in $100$ waypoints, keeping the fingers of the hand open. Then, we adaptively close the fingers in $500$ steps, maintaining the end-effector in the grasp pose. Finally, we uplift the object from the table by increasing the height of the end-effector by $2mm$ for each step, until the object is uplifted by $10cm$.

We then train our grasping policy with a modified version of the SAC~\cite{haarnoja2018soft} algorithm. Instead of starting with an empty replay buffer, we fill the initial replay buffer with demonstrations collected with the pipeline described above. We found this strategy particularly effective and easy to implement. It can be potentially applied to all the off-policy DRL algorithms without requiring any adaptation of the loss function to manage the distribution shift between the off-line demonstrations and the transitions acquired during training. The pseudo-code of the whole training procedure is reported in Alg. 1.

\section{EXPERIMENTAL SETUP}
\label{sec:exp_setup}
\subsection{Simulated Environment}
To learn our task, we deployed a simulated environment for manipulation tasks with the iCub humanoid using the MuJoCo~\cite{todorov2012mujoco} simulator (see Fig.~\ref{fig:icub_sim}). We implemented our environment as a Gym~\cite{1606.01540} interface to train policies with the main libraries implementing state-of-the-art DRL algorithms (e.g. Stable-Baselines3~\cite{stable-baselines3}).

We designed the model of the simulated iCub to be as similar as possible to the real robot, to reduce the sim-to-real gap. 
For example, in our environment, we simulate the \textit{Intel(R) RealSense D415} headset mounted on the real iCub robot that acquires RGB and depth images. Moreover, we implemented fingers actuation as in the real robot, e.g. using a tendon to simulate the actuator that controls the six joints in the little and ring fingers.

Finally, we integrated the iKin~\cite{pattacini2010experimental} and iDynTree~\cite{10.3389/frobt.2015.00006} libraries in our environment for inverse kinematics computation, and we adapted the YCB-Video~\cite{xiang2018posecnn} objects models to perform manipulation tasks and benchmark our results.

\subsection{Training Hyperparameters}

We train our policy with the implementation of the SAC~\cite{haarnoja2018soft} algorithm available in the Stable-Baselines3~\cite{stable-baselines3} library. We report the considered training hyperparameters in Tab.~\ref{tab:sac_params}. While most of the parameters are the same as the ones in the original SAC~\cite{haarnoja2018soft} implementation, we increase the number of hidden layers units to $1024$ to deal with the high-dimensionality of our state space and we increase the training frequency to $10$ timesteps to avoid implicit underparametrization~\cite{kumar2021implicit}.

\begin{table}[]
    \vspace{2mm}
    \centering
    \begin{tabular}{c|c}
        Parameter                               & Value \\
        \hline
        Optimizer                               & Adam~\cite{kingma:adam} \\
        Learning Rate                           & $3 \cdot 10^{-4}$\\
        Discount ($\gamma$)                     & $0.99$\\
        Replay Buffer size                      & $10^6$\\
        Number of Hidden Layers (all networks)  & $2$\\
        Number of Hidden Units per Layer        & $1024$\\
        Number of Samples per Minibatch         & $256$\\
        Entropy Target                          & $-15$\\
        Non-linearity                           & ReLU\\
        Target Smoothing Coefficient ($\tau$)   & $0.005$\\
        Target Update Interval                  & $1$\\
        Gradient Steps                          & $1$\\                Training Frequency                      & $10$ Timesteps\\
        Total Environment Timesteps                 & $5 \cdot 10^6$\\
    \end{tabular}
    \caption{G-PAYN Training Hyperparameters}
    \label{tab:sac_params}
\end{table}

\section{RESULTS}
\label{sec:results}
\begin{figure*}
    \vspace{2mm}
    \centering
    \includegraphics[width=0.97\linewidth]{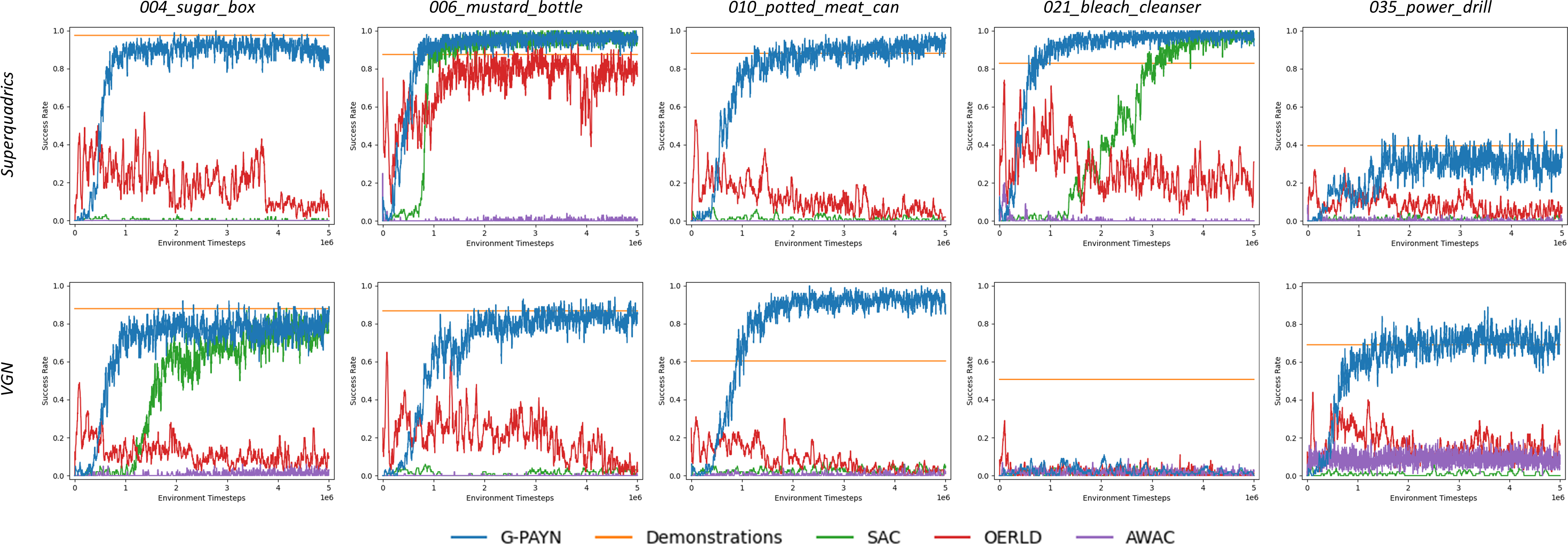}
    \caption{\textbf{Results}. We compare the considered methods for grasping execution on different objects and different grasp pose generators. In the first row we consider the approach based on superquadric modeling proposed in~\cite{vezzani2017sq} for grasp pose generation. In the second row, instead, we use VGN. In each column, we report results for different YCB-Video objects.}
    \label{fig:results}
\end{figure*}

We benchmark our approach on five objects from the YCB-Video~\cite{xiang2018posecnn} dataset. Specifically, we consider the \textit{004\_sugar\_box}, the \textit{006\_mustard\_bottle}, the \textit{010\_potted\_meat\_can}, the \textit{021\_bleach\_cleanser}, and the \textit{035\_power\_drill}. We chose these objects, that are graspable with the iCub hand, to evaluate our approach on different types of grasps. For example, when considering the superquadrics-based approach for grasp pose computation, the \textit{004\_sugar\_box} is grasped from a top-down direction, while the \textit{021\_bleach\_cleanser} is grasped from a lateral configuration. We show results in Fig.~\ref{fig:results}, where we evaluate the grasping success rate for increasing environment timesteps (see Tab.~\ref{tab:sac_params}) for different objects and grasp pose generators. For each experiment, we set the object in a random initial position that is reachable by the iCub, and we randomly rotate it around the axis perpendicular to the table.

\subsection{Baselines}
We compare the proposed \textbf{G-PAYN} (blue line in Fig.~\ref{fig:results}) to four different baselines:
\begin{itemize}
    \item \textbf{Demonstrations Pipeline} (orange line in Fig.~\ref{fig:results}): this is the approach described in Sec.~\ref{subsec:policy_training} that is used to automatically collect the demonstrations for the training. 
    Since this method does not require training, it has a constant success rate. We consider the success rate obtained while acquiring demonstrations to fill the replay buffer for the corresponding experiment.
    \item \textbf{SAC} (green line in Fig.~\ref{fig:results}): this is the standard SAC~\cite{haarnoja2018soft} algorithm trained with the hyperparameters in Tab.~\ref{tab:sac_params}. Differently from \textbf{G-PAYN}, it starts with an empty replay buffer.
    \item \textbf{OERLD} (red line in Fig.~\ref{fig:results}): we train a grasping policy with the loss function proposed in~\cite{nair2018overcoming}. For a fair comparison with the other methods, differently from the approach in~\cite{nair2018overcoming}, we combine the behavior cloning loss with the actor loss of the SAC~\cite{haarnoja2018soft} algorithm. Moreover, since we designed our task with a dense reward function with a well defined goal state, we do not apply the procedure to overcome the problem of sparse rewards described in~\cite{nair2018overcoming}, and called \textit{resets to demonstration states} by the authors. As for the other approaches, we consider as the demonstrations replay buffer the same replay buffer used at the beginning of the \textbf{G-PAYN} training. At every training step, we sample $256$ transitions from the replay buffer and $32$ transitions from the demonstrations.
    \item \textbf{AWAC} (purple line in Fig.~\ref{fig:results}): this is the approach proposed in~\cite{nair2020awac}. We rely on the implementation in~\cite{d3rlpy}. For training, we consider the default hyperparameters, setting the batch size to $256$ and the number of per layer hidden units to $1024$. During off-line training, we use as demonstrations data the same replay buffer used for \textbf{G-PAYN}'s warm start. Fig.~\ref{fig:results} reports the success rate for increasing timesteps of the on-line training stage.
\end{itemize}

\begin{figure*}
    \centering
    \includegraphics[width=0.97\linewidth]{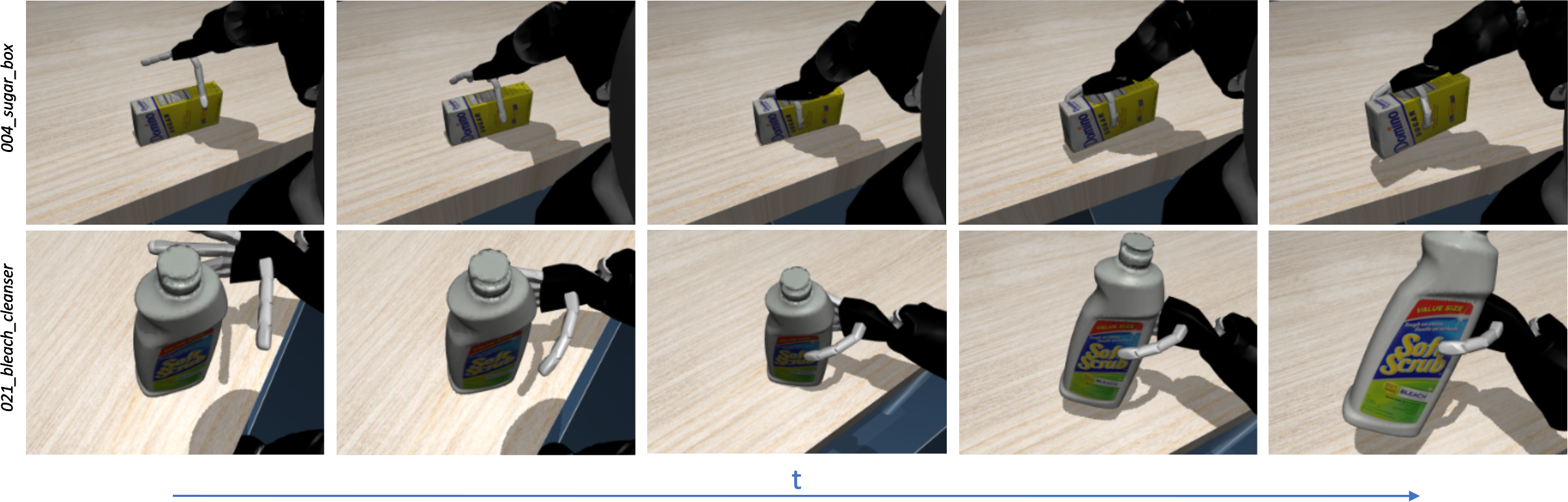}
    \caption{\textbf{Qualitative evaluation}. We show examples of our method on two of the five objects considered in the experiments (the \textit{004\_sugar\_box} and the \textit{021\_bleach\_cleanser}) using the approach based on superquadrics for grasp pose generation. We show that the learned policies manage to successfully approach the objects, grasp and uplift them.}
    \label{fig:qualitative}
\end{figure*}

\subsection{Discussion}
Results in Fig.~\ref{fig:results} show that \textbf{G-PAYN} achieves at least a comparable success rate to the \textbf{Demonstrations Pipeline}, with the exception of the \textit{021\_bleach\_cleanser} experiment when using VGN as grasp pose generator. More importantly, in half of the experiments \textbf{G-PAYN} surpasses \textbf{Demonstrations Pipeline}, and in some cases, e.g. in the \textit{010\_potted\_meat\_can+VGN} or in the \textit{021\_bleach\_cleanser+Superquadrics} experiments, its success rate outperforms the baseline by a large margin ($\sim0.3$ and $\sim0.15$ gap in the success rate for the two experiments, respectively). Notably, \textbf{G-PAYN} outplays all the DRL baselines. \textbf{SAC}, due to the high dimensionality of the problem, suffers from the lack of initial demonstrations in the replay buffer and achieves a comparable success rate to \textbf{G-PAYN} in only three experiments. Moreover, in these three cases, \textbf{SAC} requires a significantly higher number of training timesteps to achieve the same success rate as \textbf{G-PAYN}. 
\textbf{OERLD}, instead, achieves an acceptable success rate only in the \textit{006\_mustard\_bottle+Superquadrics} experiment. In all the other cases, despite the initial highest success rate obtained thanks to the behavior cloning component in the loss function, this method is not able to improve the performance throughout the training. Finally, \textbf{AWAC} achieves the worst results among the considered methods. In preliminary experiments, we noticed that training a policy on the task at hand with SAC~\cite{haarnoja2018soft} or behavior cloning on off-line data is impractical. This motivates the low success rate achieved by \textbf{AWAC}, since at the beginning of the optimization it is trained off-line on the demonstrations.

Results in Fig.~\ref{fig:results} show also the importance of the initial grasp pose used as a prior information by all the considered methods. Experiments with the \textit{021\_bleach\_cleanser} provide evidence of this. In fact, the \textit{021\_bleach\_cleanser+Superquadrics} policy achieves a success rate close to $1$, while the success rate of the \textit{021\_bleach\_cleanser+VGN} experiment is close to $0$. This is due to the fact that, for the \textit{021\_bleach\_cleanser}, grasp poses generated by the superquadrics-based algorithm are always feasible for the iCub (see also the qualitative evaluation in Fig.~\ref{fig:qualitative}), while VGN predicts top-down grasp poses which, especially for tall objects, are difficult to reach by the iCub, and add a degree of complexity to the task, requiring a much more precise grasping procedure. This aspect further motivates the need for the robotics community to develop methods that strive for both accurate grasp poses synthesis and grasp execution, as we aim to do with the proposed \textbf{G-PAYN}.

Qualitative results reported in Fig.~\ref{fig:qualitative} show that policies obtained with \textbf{G-PAYN} behave differently from the ones collected with the \textbf{Demonstrations Pipeline}. Indeed, instead of splitting the grasp execution in three phases as explained in Alg. 1, \textbf{G-PAYN} policies perform a continuous movement that closes the fingers already while approaching the object. Moreover, the trained policies manage to solve the task in fewer steps ($\sim 100$ steps for \textbf{G-PAYN} vs. $\sim 650$ steps for the \textbf{Demonstrations Pipeline}). This indicates that \textbf{G-PAYN}, while benefiting from the demonstrations as a warm start for the training, is able to learn policies that effectively optimize the task-specific reward function, instead of just imitating the off-line data. This further supports our choice to use demonstrations only as a warm start for policy training in contrast to those methods that are instead constrained to off-line data such as \textbf{OERLD} and \textbf{AWAC}. Indeed, the former attempts at imitating the off-line data throughout the entire training session with the loss component for behavior cloning. The latter, instead, learns a policy only from the demonstrations during the off-line training phase. This aspect leads to poor results for both methods, compared to the performance of our approach.

Finally, we qualitatively evaluate the \textit{006\_mustard\_bottle +Superquadrics} policy trained in simulation on the real iCub\footref{video_fn}, without any fine-tuning. While the deployment of a sim-to-real method for transferring policies on the real robot is out of the scope of this paper and left as future work, we show that our policies can be deployed on the real robot without requiring any adaptation of action and state spaces.

\section{CONCLUSIONS}
\label{sec:conclusions}
Multi-fingered grasping is an important task for robots that need to perform dexterous manipulation tasks. However, due to the difficulty of designing grasping strategies to control robotic hands with tens of DoFs, solving this task is still an open problem.

To fill this gap, we propose \textbf{G-PAYN}, a DRL approach that leverages on automatically collected demonstrations and on an initial grasp pose generated by an external algorithm for grasp synthesis. We learn the task using visual, tactile and proprioceptive information as inputs, and we show that our approach outperforms all the considered DRL baselines. Our approach also outperforms the success rate achieved by the pipeline for collection of off-line demonstrations in half of the experiments, achieving a comparable performance in almost all the remaining instances.

With our experiments, we highlight the importance of a suitable initial grasp pose to effectively learn the task. As a future work, we plan to improve our approach by developing a learning method that integrates different approaches to generate initial grasp poses and then selects the best one based on past experience.

While our approach has shown its effectiveness to learn the multi-fingered grasping task, we plan to further improve our method by speeding-up the training procedure and making it feasible on the real robot. At this aim it could be beneficial to incorporate a component for behavior cloning in the loss function during the initial learning steps, to take advantage of the steep learning curve demonstrated by \textbf{OERLD}. Finally, we plan to extend our approach to deal with distractors in the environment and with external forces.



\section*{ACKNOWLEDGMENT}

We acknowledge financial support from the PNRR MUR project PE0000013-FAIR. L. R. acknowledges the financial support of the Center for Brains, Minds and Machines (CBMM), funded by NSF STC award CCF-1231216, the European Research Council (grant SLING 819789), the AFOSR projects FA9550-18-1-7009, FA9550-17-1-0390 and BAA-AFRL-AFOSR-2016-0007 (European Office of Aerospace Research and Development), and the EU H2020-MSCA-RISE project NoMADS - DLV-777826.


\bibliographystyle{unsrt}
\bibliography{bibliography}

\end{document}